\documentclass[sigconf, nonacm]{acmart}

\setcopyright{none}
\renewcommand\footnotetextcopyrightpermission[1]{}

\AtBeginDocument{%
  }

\usepackage{bm}
\usepackage{float}
\usepackage{booktabs}
\usepackage{algorithm}
\usepackage[noend]{algpseudocode}
\usepackage{microtype}

\newcommand{\methodFullName}{Action Sequence Crossover with performance-Informed Interpolation for MAP-Elites}
\newcommand{\methodName}{ASCII-ME}
\newcommand{\variationName}{ASCII}
\newcommand{\variationFullName}{Action Sequence Crossover with performance-Informed Interpolation}
\newcommand{\isoline}{Iso+LineDD}
\newcommand{\pga}{PGA-ME}
\newcommand{\dcrl}{DCRL-ME}
\newcommand{\me}{ME}

\begin{document}

\title{Scaling Policy Gradient Quality-Diversity with Massive Parallelization via Behavioral Variations}

\author{Konstantinos Mitsides}
\email{konstantinos.mitsides23@imperial.ac.uk}
\orcid{0009-0007-1363-9777}
\affiliation{%
  \institution{Imperial College London}
  \city{London}
  \country{United Kingdom}
}

\author{Maxence Faldor}
\email{m.faldor22@imperial.ac.uk}
\orcid{0000-0003-4743-9494}
\affiliation{%
  \institution{Imperial College London}
  \city{London}
  \country{United Kingdom}
}

\author{Antoine Cully}
\email{a.cully@imperial.ac.uk}
\orcid{0000-0002-3190-7073}
\affiliation{%
  \institution{Imperial College London}
  \city{London}
  \country{United Kingdom}
}

\renewcommand{\shortauthors}{Mitsides et al.}

\begin{abstract}
Quality-Diversity optimization comprises a family of evolutionary algorithms aimed at generating a collection of diverse and high-performing solutions. MAP-Elites (\me{}), a notable example, is used effectively in fields like evolutionary robotics. However, the reliance of \me{} on random mutations from Genetic Algorithms limits its ability to evolve high-dimensional solutions. Methods proposed to overcome this include using gradient-based operators like policy gradients or natural evolution strategies. While successful at scaling \me{} for neuroevolution, these methods often suffer from slow training speeds, or difficulties in scaling with massive parallelization due to high computational demands or reliance on centralized actor-critic training. In this work, we introduce a fast, sample-efficient \me{} based algorithm capable of scaling up with massive parallelization, significantly reducing runtimes without compromising performance. Our method, \methodName{}, unlike existing policy gradient quality-diversity methods, does not rely on centralized actor-critic training. It performs behavioral variations based on time step performance metrics and maps these variations to solutions using policy gradients. Our experiments show that \methodName{} can generate a diverse collection of high-performing deep neural network policies in less than 250 seconds on a single GPU. Additionally, it operates on average, five times faster than state-of-the-art algorithms while still maintaining competitive sample efficiency.
\end{abstract}

\maketitle
\pagestyle{plain}

\begin{figure}[t]
\centering
\includegraphics[width=\linewidth]{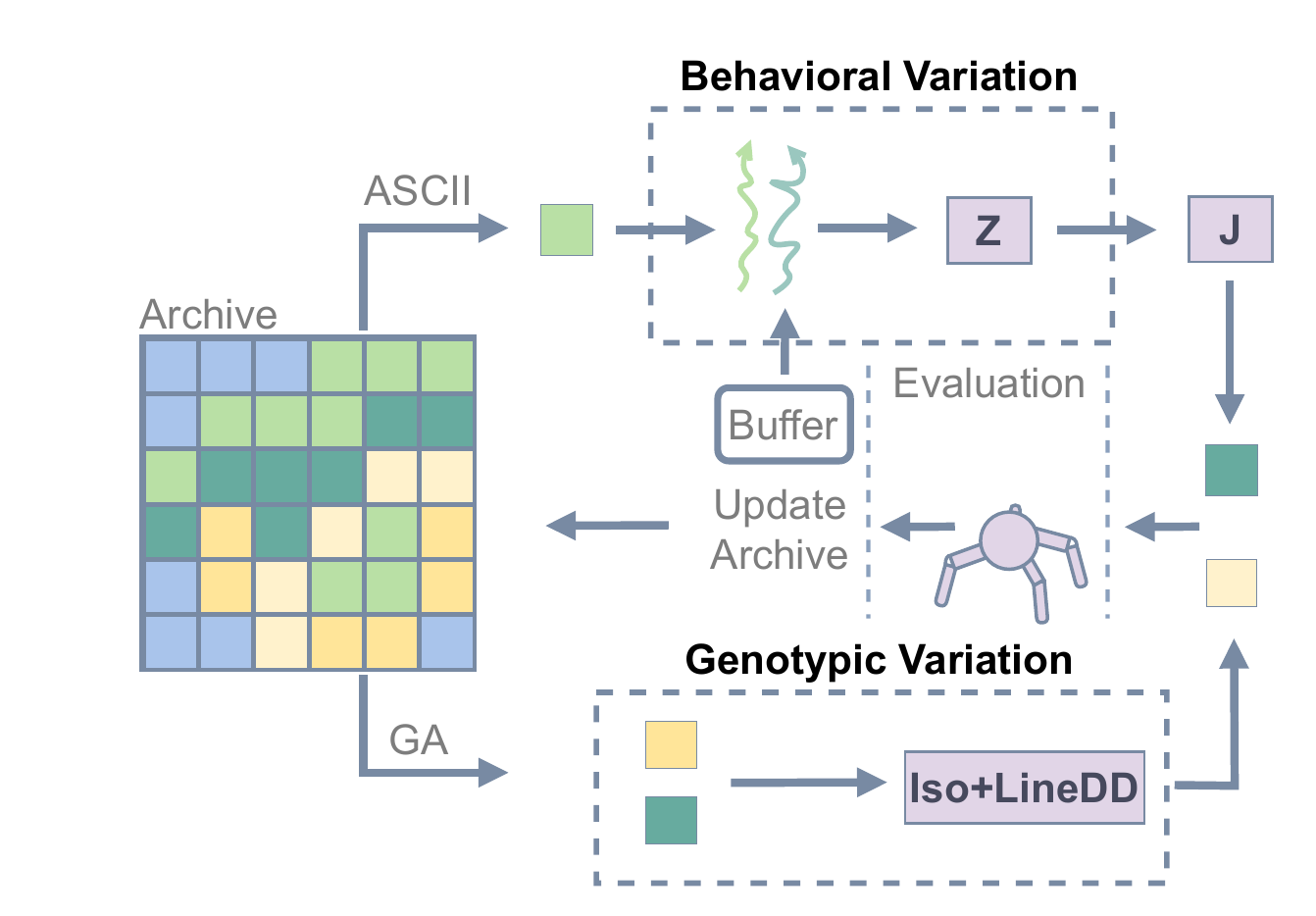}
\caption{The \methodName{} method employs two distinct variation operators within the standard MAP-Elites loop: (1) \isoline{}, which mutates a parent genotype based on that of a randomly selected elite; (2)  \variationName{}, which interpolates between the parent's behavior and another behavior sampled from the buffer, using performance metrics encapsulated in $\bm{Z}$. The behavioral changes are then mapped to the genotypic space by $\bm{J}$ to mutate the parent genotype.}
\label{fig:Method}
\end{figure}

\section{Introduction}
Quality-Diversity (QD) optimization is a family of evolutionary algorithms designed to generate a collection of solutions that are diverse and high-performing~\cite{QD-optimization, QD-framework, QD-frontier}. Unlike traditional optimization methods that focus on identifying a single optimal solution, QD optimization encourages local competition among groups of solutions with similar characteristics to identify the fittest solution within each group. By maintaining a diverse collection of solutions --- acting as ``stepping stones'' --- QD avoids local optima and explores multiple search space regions. This approach reduces the risk of premature convergence, facilitates the discovery of innovative and globally optimal solutions that traditional methods often miss~\cite{exploration1, exploration2, stepping_stones1, ME}, and enables rapid adaptation to unforeseen downstream tasks~\cite{robots_adapt_animals}. In general, QD has shown benefits across a variety of applications, from robotics~\cite{robots_adapt_animals} to design~\cite{design} and content generation~\cite{content_gen}.

A prominent algorithm in QD is MAP-Elites (\me{}) \cite{ME}, which is a straightforward yet potent and versatile method~\cite{game_content_gen, aerodynamics}. Recently, it has been employed to generate diverse adversarial prompts~\cite{rainbow_teaming}, significantly enhancing the adversarial resilience of large language models, such as Llama 3.1~\cite{llama3}. \me{} utilizes Genetic Algorithms (GAs) and typically employs the Directional Variation (\isoline{}) operator to apply random mutations to solutions, facilitating navigation within the search space. Notably, although conducting more solution evaluations in parallel reduces the total number of iterations when the evaluation budget is fixed, \me{}’s performance remains stable and even drastically reduces its runtime, thereby showcasing strong scalability~\cite{accelerated_me}. Nevertheless, its reliance on genetic variation operators lead to slow convergence in expansive search spaces, making it less effective for evolving large deep neural networks (DNNs)~\cite{robots_adapt_animals, QD-framework, ME}. 

Methods proposed to overcome this limitation include replacing or augmenting the mutation operators of \me{} with other gradient-based variation operators, such as policy gradients (PG)~\cite{Williams_1992,ac_sutton} or natural gradients in the form of evolution strategies (ES)~\cite{Daan_2011, OpenAI-ES}. ES-based \me{} algorithms have successfully scaled \me{} to neuroevolution. However, their reliance on numerous samples to estimate gradients renders them sample-inefficient and computationally demanding, limiting their practicality compared to PG methods~\cite{MEES, MEMES, OpenAI-ES}. Conversely, PG-based \me{} algorithms integrate \me{} with Deep Reinforcement Learning (DRL)~\cite{mnih_playing_2013,mnih_human-level_2015} by using PGs from a centralized actor-critic (AC) method~\cite{ac_algos,ac_sutton} to mutate solutions. While this AC training has helped \me{} achieve state-of-the-art performance in QD-RL tasks~\cite{QD_benchmark} in terms of sample efficiency~\cite{PGA, DCRL, DCG}, it undermines \me{}'s potential for speed and scalability. AC training is constrained by its reliance on numerous sequential iterations necessary for convergence, which inherently slows down performance enhancement. Moreover, within a fixed evaluation budget, the necessity to increase parallel evaluations consequently reduces the total number of iterations possible, thus limiting the training’s overall effectiveness and scalability.

In this work, we aim to efficiently evolve a collection of diverse high-performing large DNN policies, while ensuring rapid execution and scalability with massive parallelization. Specifically, we make the following contributions: \textbf{(1)} We introduce a PG-based variation operator, \variationName{}, unique among PG-based QD methods as it does not rely on centralized AC training. Utilizing the Markov Decision Process (MDP) framework, it interpolates between solution behaviors based on time step performance metrics to generate new desired behaviors. This mirrors the \isoline{} approach but operates in the behavior space, allowing efficient biasing of variations. Employing PGs, it then maps these behavioral changes into the solution space to mutate the solutions. \textbf{(2)} We integrate \variationName{} with \me{} to develop \methodName{}, which uses both the PG-based operator, \variationName{}, and the genetic operator, \isoline{}, to mutate solutions (see Figure~\ref{fig:Method}). \textbf{(3)} We compare our method’s sample and runtime efficiency across five continuous control locomotion tasks, demonstrating that it consistently outperforms established baselines in balancing these efficiencies. \textbf{(4)} We conduct an extensive analysis of how parallel evaluations affect each baseline’s performance and runtime under a fixed evaluation budget. Our experimental results show that \methodName{} robustly scales up with an increasing number of parallel evaluations, significantly decreasing runtimes without compromising performance. \textbf{(5)} We evaluate the synergy between our operator, \variationName{}, and \isoline{}, as well as their contribution to enhancing the collection of solutions in terms of both diversity and performance.

To the best of our knowledge, \methodName{} is the first PG-based QD algorithm that does not rely on AC methods, while still capable of evolving DNN policies with thousands of parameters at competitive sample and runtime efficiency. This, combined with its strong scalability on a single GPU, underscores the potential of this promising new framework for non-AC PG-based QD methods.

\section{Background \& Related Work}
\subsection{Problem Setting}
\label{subsec:prob_setting}
We consider an agent interacting with an environment at discrete time steps $t$ for an episode of length $H$. At each time step $t$, the agent observes a state $\bm{s}_t$, executes an action $\bm{a}_t$ and receives a scalar reward $r_{t+1}$. We model this task as a Markov Decision Process (MDP) that includes a state space $\mathcal{S}$, a continuous action space $\mathcal{A}$, a stationary transition dynamics distribution $p(\bm{s}_{t+1}|\bm{s}_t, \bm{a}_t)$ and a reward function $r: \mathcal{S}\times \mathcal{A}\rightarrow \mathbb{R}$. In this work, the policy is deterministic and represented by a neural network parameterized by a genotype $\bm{x} \in \mathcal{X}$ and denoted $\bm{\mu}_{\bm{x}} : \mathcal{S} \rightarrow \mathcal{A}$. Using its policy, the agent interacts with the environment for an episode to generate a trajectory consisting of states, actions, and rewards, denoted as $\left\{\bm{s}_t, \bm{a}_t, r_{t+1}, \bm{s}_{t+1}\right\}_{t=0}^{H-1}$. The fitness of a genotype (solution) is determined by the function $F:\mathcal{X}\rightarrow\mathbb{R}$, defined as the expected return $\mathbb{E} \left[\sum_{t=1}^{H}r_t \right]$. The objective of quality-diversity algorithms is to populate the feature space $\mathcal{D}$ with high-fitness genotypes, using a user-defined feature function $D:\mathcal{X} \rightarrow \mathcal{D}$ that categorizes solutions by the desired diversity type.

\label{sec:background}
\subsection{Quality-Diversity Optimization}
Quality-Diversity (QD) optimization~\cite{QD-optimization, QD-framework, QD-frontier} aims to generate a diverse collection of high-performing solutions to a problem by defining diversity according to a feature vector, also referred to as a set of descriptors or measures in the literature. Thus, each solution has an attributed fitness $f$, quantifying its quality; and a feature vector $\bm{d}$, describing its novelty with respect to other solutions.

\subsubsection{\textbf{MAP-Elites}}
\label{ME}
MAP-Elites (ME) is a QD algorithm that discretizes the feature space, $\mathcal{D}$, into a multi-dimensional grid of cells known as archive, and searches for the highest performing solution in each cell. \me{} starts by generating a set of $k$ random solutions that are added to the archive. The process then continues iteratively until $I$ solutions have been evaluated, following these steps: (1) a batch of $k$ solutions from the archive are uniformly selected and modified through a genetic variation operator such as mutations and/or crossovers to produce offspring, (2) evaluate the fitness and feature vector of each offspring, placing each in the cell corresponding to its feature vector if it either improves upon the current occupant's fitness or is the first to occupy that cell.

\subsubsection{\textbf{Directional Variation}}
\label{subsec:isoline}
A common and effective variation operator used in \me{} to modify the solutions is the directional variation (\isoline{}) operator~\cite{isoline_variation}. A parent, $\bm{x}_i$ is selected uniformly among the elites in the archive and the offspring $\bm{x}_i'$ is then produced using the following operator:
\begin{equation}
    \bm{x}_i' = \bm{x}_i + \sigma_1 \mathcal{N}(\bm{0},\bm{I}) + \sigma_2 (\bm{x}_j - \bm{x}_i) \mathcal{N}(0,1)
\end{equation}

In other words, $\bm{x}_i'$ is the resultant vector of $\bm{x}_i$, and another two vectors. The first vector is a randomly generated vector, while the second vector aligns with the direction of correlation between $\bm{x}_i$ and a different randomly chosen elite $\bm{x}_j$. The direction of the second vector --- towards or away from $\bm{x}_j$ --- is determined by a randomly generated scalar constant, and the magnitude of the vector is proportional to the Euclidean distance between $\bm{x}_i$ and $\bm{x}_j$, $\big|\big|\bm{x}_j-\bm{x}_i\big|\big|_{2}$.

\subsubsection{\textbf{Accelerating MAP-Elites with Parallelization}}
Recent advancements in hardware acceleration, such as GPUs and TPUs, have introduced tools that parallelize task processing, significantly reducing the runtime of many optimization algorithms. Highly-parallel simulators like Brax~\cite{brax} and Isaac~\cite{isaac} have enhanced robotic simulations, leading to the development of new QD libraries like QDax~\cite{qdax}. Using Brax, QDax can perform 10 to 100 times more evaluations per iteration in parallel within the same timeframe compared to traditional QD algorithms that rely on CPU-based simulations. Lim et al.~\cite{accelerated_me} demonstrated that, although conducting more evaluations in parallel reduces the number of iterations when the evaluation budget is fixed, \me{}’s performance remains stable while its runtime is reduced by up to approximately $100$ times.

\subsection{Deep Reinforcement Learning}
\label{subsec:DRL}
Deep Reinforcement Learning (DRL)~\cite{mnih_playing_2013,mnih_human-level_2015} combines reinforcement learning (RL) with deep neural networks (DNNs) to approximate functions, representing policies and value functions in high-dimensional spaces. Unlike black-box optimization methods like evolutionary algorithms, DRL leverages the Markov Decision Process (MDP) framework to enhance sample efficiency by utilizing time step information.

Policy gradient (PG) methods is a family of DRL algorithms that aim to optimize the fitness function $F(\bm{x})$ --- find the genotype $\bm{x^{*}}$ which maximizes F --- by adjusting the genotype in the direction of the estimated fitness gradient:
\begin{equation}
    \bm{x}_{k+1}=\bm{x}_{k}+ \alpha\widehat{\nabla F(\bm{x}_{k})}
\end{equation}
where traditionally,
\begin{equation}
    F(\bm{x}):= \mathbb{E}_{\pi_{\bm{x}}} \left[G_0 \right]
\end{equation}
represents the expected reward-to-go at the initial time step of policy $\pi_{\bm{x}}$. The reward-to-go at time step $t$ of an episode of length $H$ is defined as: 
\begin{equation}
    G_t=\sum_{h=t}^{H-1}\gamma^{h-t}r_{h+1}.
\end{equation}

Building on PG methods, actor-critic (AC)~\cite{ac_sutton, ac_algos} algorithms divide the optimization process into two interconnected components: the actor, which updates the policy parameters (genotype), and the critic, which estimates the value functions to guide the actor’s updates. As the learning progresses through sequential steps, the critic’s value estimates become increasingly accurate, providing the actor with more reliable information for gradient-based updates.

\subsection{Related Work}
The challenge of evolving diverse solutions in a high-dimensional search space has recently been prominent. Recent QD approaches, based on the \me{} method, emulate its strategy for maintaining solutions and using a discretized multidimensional archive. However, they diverge in their design of more efficient variation operators, and some also differ in how solutions are sampled from the archive. We categorize these algorithms into three groups: those using the TD3~\cite{TD3}, the OpenAI-ES~\cite{OpenAI-ES}, and the CMA-ES~\cite{CMA-ES} algorithms for sampling and variation. Note that TD3 and OpenAI-ES have been applied in recent third-group variants to scale QD to neuroevolution.

\subsubsection{\textbf{TD3 based}}
\label{subsec:ac_based_methods}
\pga{}~\cite{PGA} extends \me{} by incorporating an additional PG-based operator. It utilizes an AC method (TD3), using transitions from evaluations saved in a replay buffer to maximize an objective that is solely based on fitness. Using the trained critic, a portion of the selected solutions from the archive are mutated, while the rest are mutated using the \isoline{} variation operator. Additionally, the greedily trained actor from the AC training is also considered for inclusion in the archive (Actor Injection). \dcrl{}~\cite{DCRL} builds on PGA-ME by adapting its AC training to be conditioned on the feature vectors (descriptors) of the solutions. Consequently, a descriptor-conditioned critic is employed to mutate the solutions. In addition to the mutated solutions, multiple trained actors, each conditioned on randomly selected descriptors, are considered for inclusion in the archive. QD-PG~\cite{QD-PG} replaces the GA variation operator in \me{} with one based on PGs. It collects transitions from evaluations, employing an AC training method (TD3) that maximizes objective functions based on diversity and fitness. Using the trained diversity and fitness critics, it then mutates the solutions for addition to the archive. The PG-based \me{} methods significantly improve the sample efficiency of QD optimization, evolving neural networks with thousands of parameters. 

However, they require extensive time for AC training, which considerably slows down the process. Additionally, the sequential nature of AC training, necessary for convergence, limits the benefits from parallelization, thereby constraining their scalability. 

\subsubsection{\textbf{OpenAI-ES based}}
ME-ES~\cite{MEES} replaces the GA variation operator in \me{} with ES. It sequentially samples one solution randomly from the archive and uses the OpenAI-ES algorithm to optimize it by alternating between maximizing a novelty function and a fitness function. The gradients for either fitness or novelty are locally estimated from multiple perturbed instances of a parent solution, generating a new solution. MEMES~\cite{MEMES} scales up ME-ES by eliminating the sequential component, thereby enabling independent ES processes to optimize multiple solutions in parallel. A portion of these processes is dedicated to optimizing for diversity, while the rest focuses on fitness. Although MEMES employs black-box optimization, it demonstrates competitive performance against PG-based QD algorithms in QD-RL tasks, when a sufficient evaluation budget is available. 

While MEMES theoretically offers robust scalability --- since optimizing additional solutions in parallel can only enhance its performance --- it requires significant computational resources, which may not be feasible on standard consumer hardware. 

\subsubsection{\textbf{CMA-ES based}}
CMA-ES~\cite{CMA-ES} has been integrated with \me{} to form CMA-ME~\cite{CMA-ME}. Unlike \me{}, which applies mutations to randomly selected solutions from the archive, CMA-ME strategically uses CMA-ES to mutate solutions, focusing on intrinsic objectives to enhance archive quality. Since its inception, many new algorithms that build on and improve CMA-ME have been developed~\cite{MAE, scale_MAE, scale_MAEGA}. Among these, PPGA~\cite{PPGA} stands out as the state-of-the-art algorithm operating under the Differential Quality Diversity (DQD)~\cite{DQD} framework. It integrates Proximal Policy Optimization (PPO)~\cite{PPO} with CMA-MAEGA~\cite{MAE} and alternates between estimating fitness and feature gradients using PPO and maintaining a population of coefficients. These coefficients are optimized to maximize archive improvement by linearly combining the gradients.

PPGA achieves competitive results compared to state-of-the-art QD algorithms for QD-RL tasks; however, its requirement for a significant number of evaluations makes it sample inefficient.

\section{Methods}
\label{sec:methods}
In this work, we introduce the \methodFullName{} (\methodName{}) algorithm (see Appendix~\ref{apdx:pseudocode}). It is an extension of \me{} that targets evolving DNN policies while --- unlike other extensions --- retaining the speed and scalability of ME. Following the standard \me{} loop, a portion of the solutions uniformly sampled from the archive is processed using the \isoline{} operator, while the remainder are processed by our new \variationFullName{} (\variationName{}) operator (see Figure~\ref{fig:Method}). This operator creates action-based variations by interpolating between two action sequences based on time step performance metrics and applies PGs to modify the solutions. Subsequently, all processed solutions from both operators are evaluated for one episode to determine their eligibility for inclusion in the archive, while the evaluation information is stored in a buffer for use in the next iteration. One action sequence is associated with a sampled solution from the archive, and the other is derived from information in the buffer, which is linked to the evaluation of a previously mutated solution. A key distinction between \methodName{} and other PG-based QD algorithms is that \methodName{} does not utilize a centralized AC training, making the algorithm faster and more scalable with massive parallelization.

\subsection{\variationName{} Operator}
\label{subsec:variation_eq}
Inspired by the \isoline{} variation operator, as detailed in Section~\ref{subsec:isoline}, \variationName{} samples two action sequences and, based on their time step performance, interpolates between them to generate a new sequence. This sequence is then used to define the gradient step for policy updates through back-propagation. Specifically, \variationName{} iteratively applies the following variations to the sampled genotype $\bm{x}_i$ over $e$ iterations to produce its offspring:
\begin{equation}
    \bm{A}_{\bm{x}_i}' = \bm{A}_{\bm{x}_i} + \lambda_1 \mathcal{N}(\bm{0}, \bm{I}) + \lambda_2 \bm{Z}_{\bm{x}_i,\bm{x}_j} \left[\bm{A}_{\bm{x}_j} - \bm{A}_{\bm{x}_i}\right]
\label{eq:action_based_variation}
\end{equation}
\begin{equation}
    \bm{x}_i' = \bm{x}_i + \bm{J}_{\bm{x}_i}^{T}\Delta\bm{A}_{\bm{x}_i},
\label{eq:parameter_based_mutation}
\end{equation}
where $\Delta\bm{A}_{\bm{x}_i}$ denotes the difference between the new and the old action sequence, reflecting the desired changes to the policy outputs. Here, $\bm{A}_{\bm{x}_i}$ and $\bm{A}_{\bm{x}_j}$ represent the action sequences consisting of time step actions for each time step in an episode, associated with the policies of genotypes $\bm{x}_i$ and $\bm{x}_j$, respectively (more details in Section~\ref{subsec:action_sequences}). The term $\bm{Z}_{\bm{x}_i,\bm{x}_j}$ is a performance-based weight matrix derived from the evaluation of corresponding genotypes (more details in Section~\ref{subsec:weight_matrix}). $\bm{J}_{\bm{x}_i}$ represents the Jacobian matrix of the policy associated with $\bm{x}_i$, mapping changes in the action space back to the genotypic space (more details in Section~\ref{subsec:jacobian}).

The variation presented by Equation~\ref{eq:action_based_variation} adopts the concept of \isoline{} operator, however, it differs in two significant ways: 1) the variation is applied to an action sequence rather than directly to the genotype; 2) a different weight is applied to each time step action difference based on the performance of the corresponding time step. These weights are encapsulated in the matrix $\bm{Z}_{\bm{x}_i, \bm{x}_j}$, contrasting with the uniform application of $\sigma_2 \mathcal{N}(0,1)$ to all parameter differences in \isoline{}. Furthermore, we set $\lambda_1 = 0$ since random exploration is already provided by \methodName{} from the \isoline{} operator in the genotypic space. The analysis of synergies between these two potential sources of random exploration --- parameter-based through \isoline{} and action-based through \variationName{} --- is deferred to future work.

\subsection{\textbf{Action Sequences and Jacobian}}
\label{subsec:notation}
Let $\bm{x}_i$ represent the genotype uniformly sampled from the archive for mutation, and $\bm{\mu}_{\bm{x}_i}$ its corresponding deterministic policy, referred to as the ``mutated policy''. The states visited by $\bm{\mu}_{\bm{x}_i}$, along with their associated rewards-to-go computed from the rewards collected during a single evaluation episode, are denoted by $\left\{\bm{s}_t^{i},G_t^{i}\right\}_{t=0}^{H-1}$. These states and rewards-to-go are also taken from the archive along with the genotype $\bm{x}_i$. Moreover, let $\left\{(\bm{s}_t^{j}, \bm{a}_t^{j}, G_{t}^{j})\right\}_{t=0}^{H-1}$ represent the information sampled from the buffer. This collection comprises the states and actions taken by $\bm{\mu}_{\bm{x}_j}$ during the evaluation phase of the previous iteration, where $\bm{x}_j$ represents a mutated genotype from that iteration. We refer to the deterministic policy associated with the genotype $\bm{x}_j$, $\bm{\mu}_{\bm{x}_j}$, as the ``target policy''. 

The buffer often includes data from genotypes not included in the archive, providing no guarantee of data quality. Despite this, it has proven more effective than using only data from the archive, likely because it introduces more diversity and promotes greater exploration of the search space (see Appendix~\ref{apdx:data_sampling}).
\subsubsection{\textbf{Action Sequences}}
\label{subsec:action_sequences}
To enable a fair comparison, we need to compare how the action sequences from the two policies differ when faced with the same states. For this, we take the state sequence of the target policy as reference, $\left\{\bm{s}_t^{j}\right\}_{t=0}^{H-1}$. Consequently, we introduce an ``imaginary'' action sequence for $\bm{x}_i$. This sequence represents the actions that the mutated policy $\bm{\mu}_{\bm{x}_i}$,  would have taken if it had followed the state sequence of the target policy $\bm{\mu}_{\bm{x}_j}$. This imaginary sequence, represented by $\bm{A}_{\bm{x}_i}$, and the actual action sequence of the target policy $\bm{\mu}_{\bm{x}_j}$, represented by $\bm{A}_{\bm{x}_j}$, are both depicted as block vectors in $\mathbb{R}^{H|\mathcal{A}|}$. Specifically:
\begin{equation}
\begin{aligned}
    \bm{A}_{\bm{x}_i} &= \begin{bmatrix}
        \tilde{\bm{a}}_0 &  \ldots & \tilde{\bm{a}}_{H-1}
    \end{bmatrix}^{T}, \quad
    \bm{A}_{\bm{x}_{j}} = \begin{bmatrix}
        \bm{a}_0 &  \ldots & \bm{a}_{H-1}
    \end{bmatrix}^{T}.
\end{aligned}
\end{equation}
where $\tilde{\bm{a}}_t=\bm{\mu}_{\bm{x}_i}(\bm{s}_t^{j})$ and $\bm{a}_t=\bm{\mu}_{\bm{x}_j}(\bm{s}_t^{j})$ are both in $\mathbb{R}^{|\mathcal{A}|}$.

\subsubsection{\textbf{Jacobian}}
\label{subsec:jacobian}
The Jacobian matrix $\bm{J}_{\bm{x}_i}\in \mathbb{R}^{H|\mathcal{A}|\times |\mathcal{X}|}$ is defined as a block matrix:
\begin{equation}
    \bm{J}_{\bm{x}_{i}}= \begin{bmatrix}
        \bm{B}_0(\bm{x}_i) & \bm{B}_1(\bm{x}_i) & \ldots & \bm{B}_{H-1}(\bm{x}_i)
    \end{bmatrix}^{T},
\end{equation}
where 
\begin{equation}
    \bm{B}_t(\bm{x}_i)=\frac{\partial\bm{\mu}_{\bm{x}}(\bm{s}^j_t)}{\partial\bm{x}}\bigg|_{\bm{x}=\bm{x}_i},
\end{equation}
and each $\bm{B}_t$ is defined in $\mathbb{R}^{|\mathcal{A}|\times |\mathcal{X}|}$.

\subsection{\textbf{Performance based Weight Matrix}}
\label{subsec:weight_matrix}
The performance based weight matrix is defined as a block matrix:
\begin{equation}
\bm{Z}_{\bm{x}_i,\bm{x}_j}=\operatorname{diag}(\bm{z}_0(\bm{x}_i, \bm{x}_j), \ldots, \bm{z}_{H-1}(\bm{x}_i, \bm{x}_j)),    
\end{equation}
where $\bm{z}_t(\bm{x}_i, \bm{x}_j)=z_t(\bm{x}_i, \bm{x}_j)\bm{I}$, with $\bm{I} \in \mathbb{R}^{|\mathcal{A}| \times |\mathcal{A}|}$. The weight $z_t(\bm{x}_i,\bm{x}_j)$ influences the interpolation between two actions at time step $t$ based on four factors: 1) the performance gain of each action, measured by comparing the rewards-to-go, as defined in Section~\ref{subsec:DRL}; 2) the similarity of the states in which each action is taken, determined by the cosine similarity function; 3) the similarity between the actions, using a Squared Exponential Kernel; and 4) a clipping mechanism to prevent large updates.

\subsubsection{\textbf{Rewards-to-go Difference}}
To compare the performance of the two action sequences at each time step, we compute the difference between the reward-to-go of the target policy, $\bm{\mu}_{\bm{x}_j}$, and the reward-to-go of the mutated policy, $\bm{\mu}_{\bm{x}_i}$, at each time-step $t$:
\begin{equation}
    \Delta G_t = G_t^{j}-G_t^{i}.
\label{eq:quality_difference}
\end{equation}

\subsubsection{\textbf{State Cosine Similarity}}
The rewards-to-go for the mutated policy, $\bm{\mu}_{\bm{x}_i}$, are computed based on a different state sequence, $\left\{\bm{s}_t^{i}\right\}_{t=0}^{T-1}$, during which it took different actions, not corresponding to those in $\bm{A}_{\bm{x}_i}$. To address this discrepancy, we adjust $\Delta G_t$ by multiplying it by the cosine similarity between the states at each time step, which quantifies how similar the states are:
\begin{equation}
\operatorname{max}\left(b,  \frac{\bm{s}_t^{i} \cdot \bm{s}_t^{j}}{||\bm{s}_t^{i}||_2 ||\bm{s}_t^{j}||_2}\right).
\label{eq:cosine_sim}
\end{equation}
The predefined constant $b$, a non-negative threshold not exceeding 1, is usually set to  $b=0.25$. By using the $\operatorname{max}$ function in conjunction with  $b$ , we ensure that all state pairs have a minimum positive weight, thus preventing weights from reaching zero or becoming negative. Zero weights would make the update matrix more sparse, whereas negative weights would reverse the sign of  $\Delta G_t$ , potentially providing misleading information about the performance of actions at each time step.
\subsubsection{\textbf{Action Kernel}}
To mitigate the inaccuracy of using the action sequence of the mutated policy, $\bm{\mu}_{\bm{x}_i}$, based on a state sequence it did not follow, we incorporate a Squared Exponential Kernel:
\begin{equation}
    k(\bm{a}_t,\tilde{\bm{a}}_t)=\exp\left\{-\frac{1}{2\sigma^{2}}\big|\big|\bm{a}_t-\tilde{\bm{a}}_t\big|\big|_2^{2}\right\}.
\end{equation}
This kernel penalizes significant deviations between actions taken in response to the same state, thereby assigning less weight to mutations involving two policies that significantly diverge from each other.

Combining these three mechanisms, we get:
\begin{equation}
    \beta_t(\bm{x}_i, \bm{x}_j) = k(\bm{a}_t,\tilde{\bm{a}}_t)\operatorname{max}\left(b,  \frac{\bm{s}_t^{i} \cdot \bm{s}_t^{j}}{|\bm{s}_t^{i}|_2 |\bm{s}_t^{j}|_2}\right)\Delta G_t.
\label{eq:unclipped_beta}
\end{equation}
Recall that $\bm{a}_t,\tilde{\bm{a}}_t, \bm{s}_t^{i}, \bm{s}_t^{j}$, and $\Delta G_t$ are all time step information collected from the mutated policy, $\bm{\mu}_{\bm{x}_i}$, and the target policy, $\bm{\mu}_{\bm{x}_j}$, thus making $\beta_t$ dependent on genotypes $\bm{x}_i$ and $\bm{x}_j$.

\subsubsection{\textbf{Proximal Clipping like in PPO}}
We further include a predefined threshold, $\epsilon$, usually set $\epsilon=0.8$, such that if the value of the kernel falls below this threshold and $\Delta G_t$ is negative, then that time step is not considered in updates:
\begin{equation}
z_t(\bm{x}_i, \bm{x}_j) = 
\begin{cases}
0 & \text{if } \ k(\bm{a}_t,\tilde{\bm{a}}_t) < \epsilon \ \ \text{and} \ \ \Delta G_t < 0, \\
\beta_t(\bm{x}_i, \bm{x}_j) & \text{otherwise}.
\end{cases}
\label{eq:pessimistic_function}
\end{equation}
This ensures that we do not degrade the performance of the mutated policy, $\bm{\mu}_{\bm{x}_i}$, by applying large updates where it is already outperforming the target policy, $\bm{\mu}_{\bm{x}_j}$, and actions deviate from each other significantly. This approach, akin to the clipping mechanism in PPO~\cite{PPO}, helps avoid excessively large mutations. 

\section{Experiments \& Results}
\begin{figure*}[ht!]
  \centering
  \includegraphics[width=\textwidth]{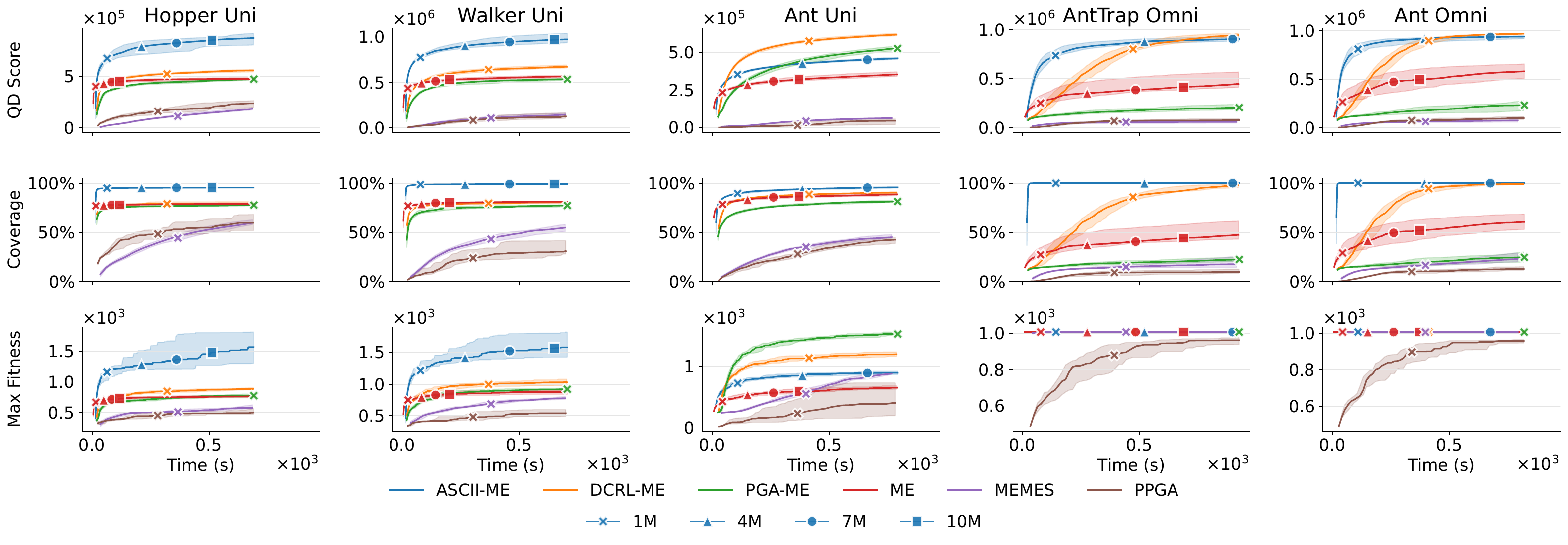}
  \caption{Main metrics for \methodName{} and baselines across tasks, with each algorithm running for the duration required for the slowest one to complete one million evaluations. The solid line represents the median, and the shaded area indicates the lower and upper quartiles across 20 seeds. Checkpoints show the number of evaluations completed by each algorithm at that point.}
  \label{fig:question1}
\end{figure*}
\label{sec:experiments}
The goal of our experimental evaluation is to answer the following three questions: \textbf{(1)} How does \methodName{} compare to AC-based QD methods in terms of sample and runtime efficiency? Moving to considerations of operational scalability, \textbf{(2)} for a fixed evaluation budget, does \methodName{} benefit from employing a larger batch size --- evaluating more solutions in parallel --- and thus fewer iterations? This question pertains to both final performance and total runtime. Further exploring the method’s components, \textbf{(3)} how well does the \variationName{} operator synergize with the \isoline{} operator, and what is the contribution of each to improving the archive? 

\subsection{Evaluation Tasks}
\setlength{\tabcolsep}{3pt}
\begin{table}[h]
\small
\caption{Evaluation Tasks}
\label{tab:tasks}
\centering
\newdimen\length
\length=1cm
\begin{tabular}{l | c c c c c}
    \toprule
    & \textsc{Ant} & \textsc{AntTrap} & \textsc{Walker} & \textsc{Hopper} & \textsc{Ant}\\
    & \includegraphics[height=0.8\length, width=\length]{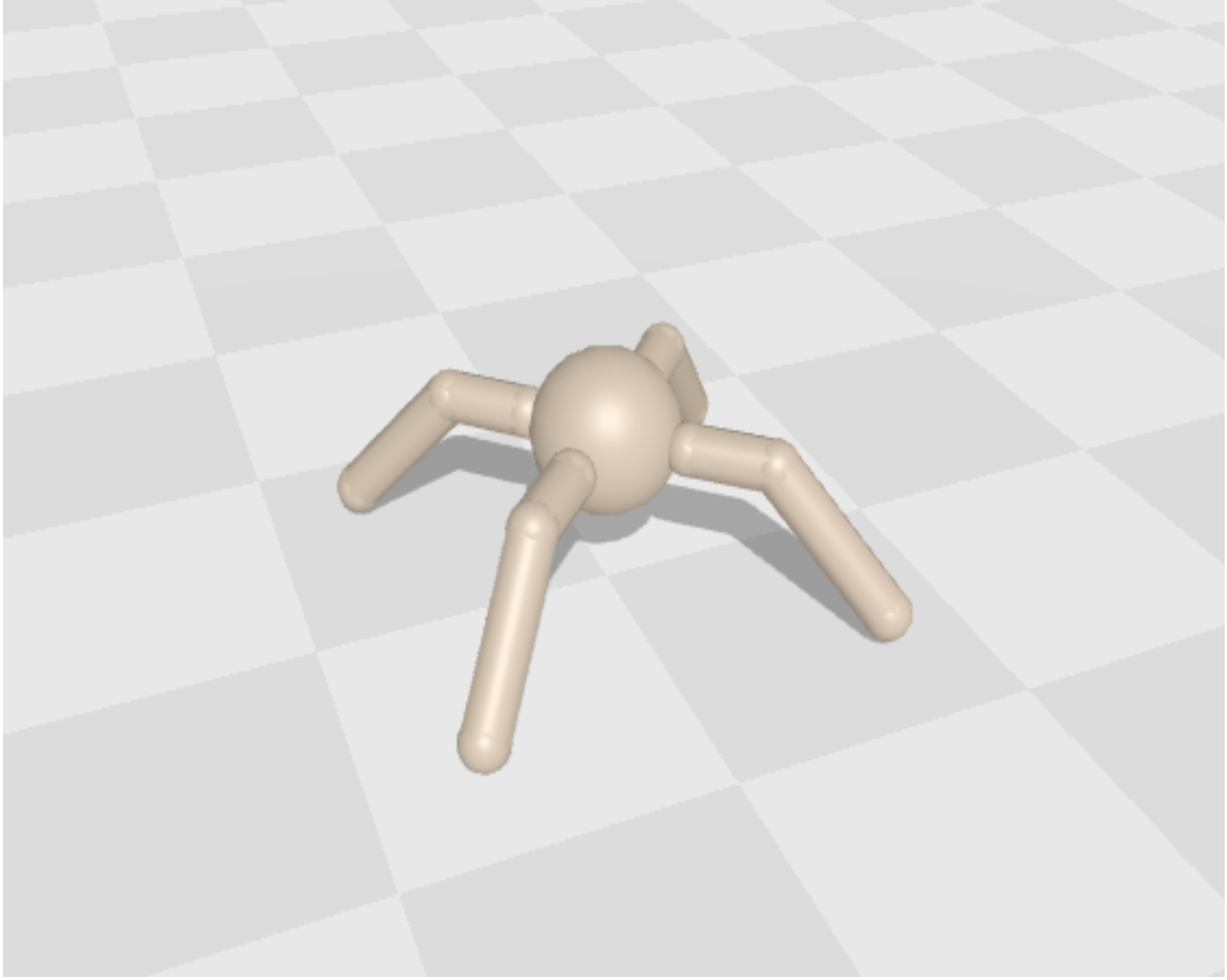} & \includegraphics[height=0.8\length, width=\length]{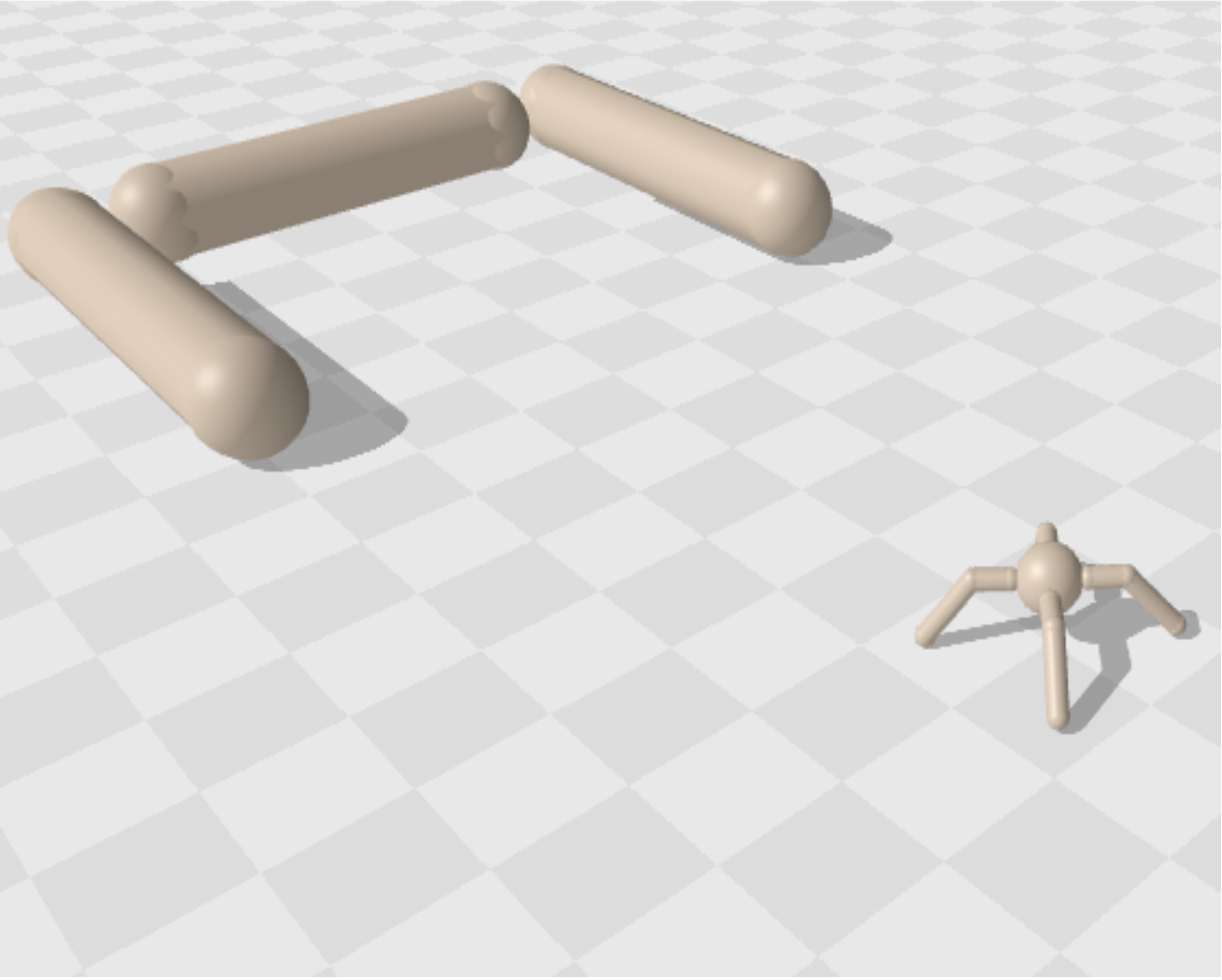} & \includegraphics[height=0.8\length, width=\length]{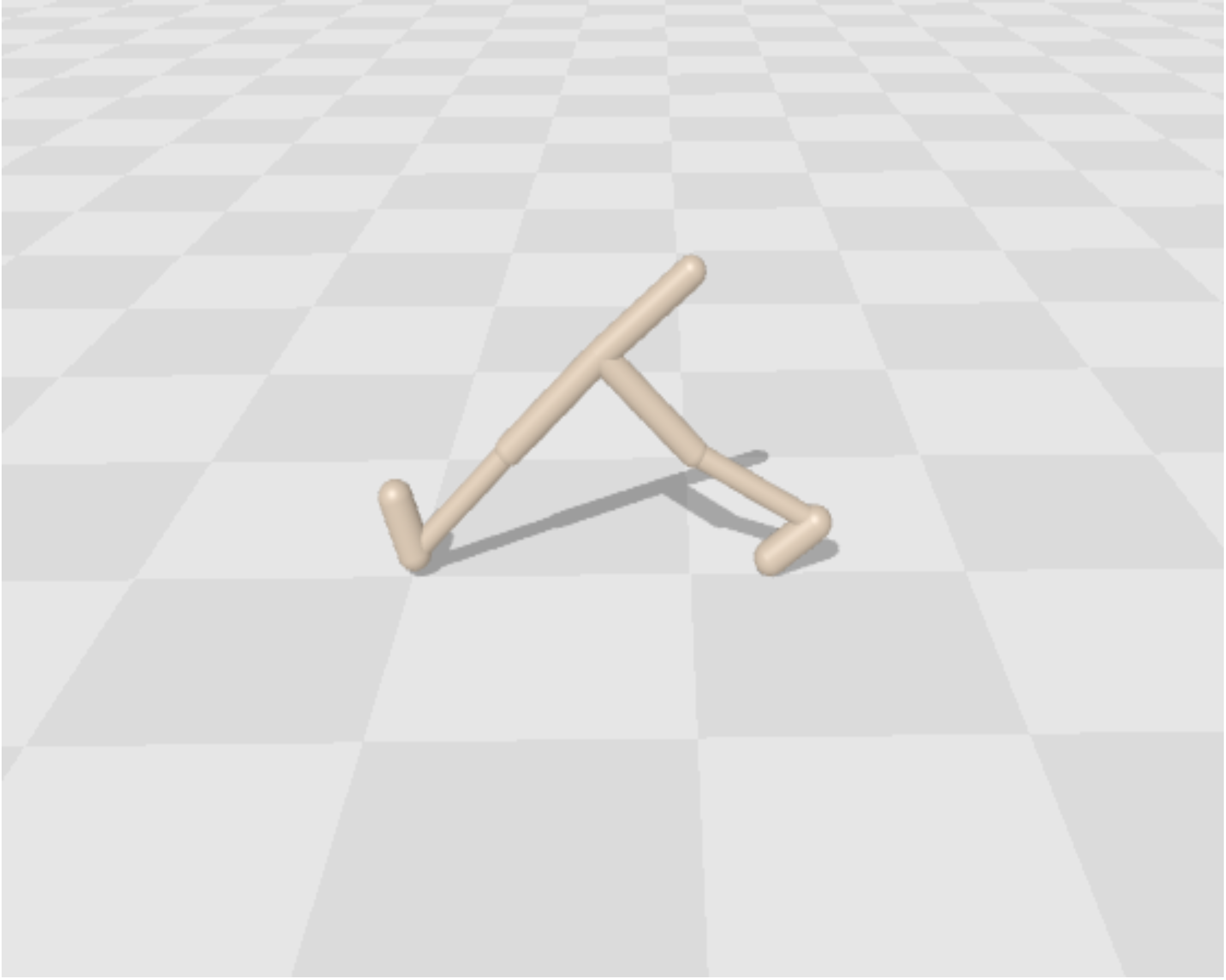} & \includegraphics[height=0.8\length, width=\length]{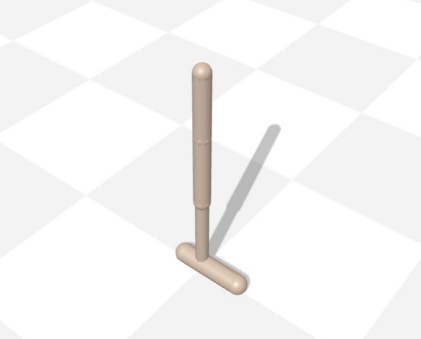} &
    \includegraphics[height=0.8\length, width=\length]{ant_uni.png}\\
    \midrule
    \textsc{State} & \multicolumn{5}{c}{Position and velocity of body and joints}\\
    \textsc{Action} & \multicolumn{5}{c}{Torques for each joint}\\
    \textsc{Type} & Omni & Omni & Uni & Uni & Uni\\
    \textsc{Descriptor} & \multicolumn{2}{c}{Final $(x, y)$ position} & \multicolumn{3}{c}{Feet contact}\\
    \textsc{State dim} & 30 & 30 & 18 & 12 & 28\\
    \textsc{Action dim} & 8 & 8 & 6 & 3 & 8\\
    \textsc{Episode len} & 250 & 250 & 250 & 250 & 250\\
    \textsc{Parameters} & 6664 & 6664 & 5766 & 5187 & 6536\\
    \bottomrule
\end{tabular}
\end{table}

We evaluate \methodName{} on five continuous control locomotion QD tasks~\cite{QD_benchmark} as implemented in Brax~\cite{brax} and derived from standard RL benchmarks~\cite{gym_envs} (see Table~\ref{tab:tasks}). AntTrap Omni and Ant Omni are \textit{omnidirectional} tasks, where the simulated robot explores a predefined two-dimensional space defined by descriptor limits, aiming to minimize energy consumption. The fitness here is determined by the accumulated reward for surviving each time step, penalized by energy usage, while the behavioral descriptor (BD) is the robot’s final xy-position. In contrast, Ant Uni, Walker Uni, and Hopper Uni are \textit{unidirectional} tasks focused on discovering diverse locomotion strategies while balancing speed against energy consumption. Fitness for these tasks is defined by the forward progress made, combined with survival time and an energy consumption penalty, while the BD is the contact rate of each foot. Task states include measurements like the center of gravity height, (x,y,z)-velocities, roll, pitch, yaw angles, and the relative position of the robot’s joints. Actions are delivered as continuous-valued torques to control the robot’s joints, with initial positions sampled from a Gaussian distribution, adding a stochastic element to the tasks.

\subsection{Baselines}
In our evaluation, we compare \methodName{} with five baseline methods, as implemented in QDAX~\cite{qdax}: \me{}~\cite{ME}, MEMES~\cite{MEMES}, \pga{}~\cite{PGA}, \dcrl~\cite{DCRL}, and PPGA~\cite{PPGA}. \me{} lays the groundwork for many advanced QD algorithms, while MEMES is a leading ES-based \me{} method. PPGA is state-of-the-art among methods utilizing CMA-ES, and for its implementation, we adapted the authors' code to suit our specific tasks. \dcrl{}, recognized as the leading QD algorithm, surpasses all others in sample efficiency for QD-RL tasks. We further include \pga{} in our comparison because, unlike \dcrl{}, it is a state-of-the-art algorithm that does not utilize descriptor information, similar to our method. Additionally, to tailor \methodName{} to our experimental needs, we conducted hyperparameter tuning, resulting in variations from \me{}'s hyperparameters, such as in the initialization of the archive.

\subsection{Metrics}
\begin{figure*}[ht!]
  \centering
  \includegraphics[width=\textwidth]{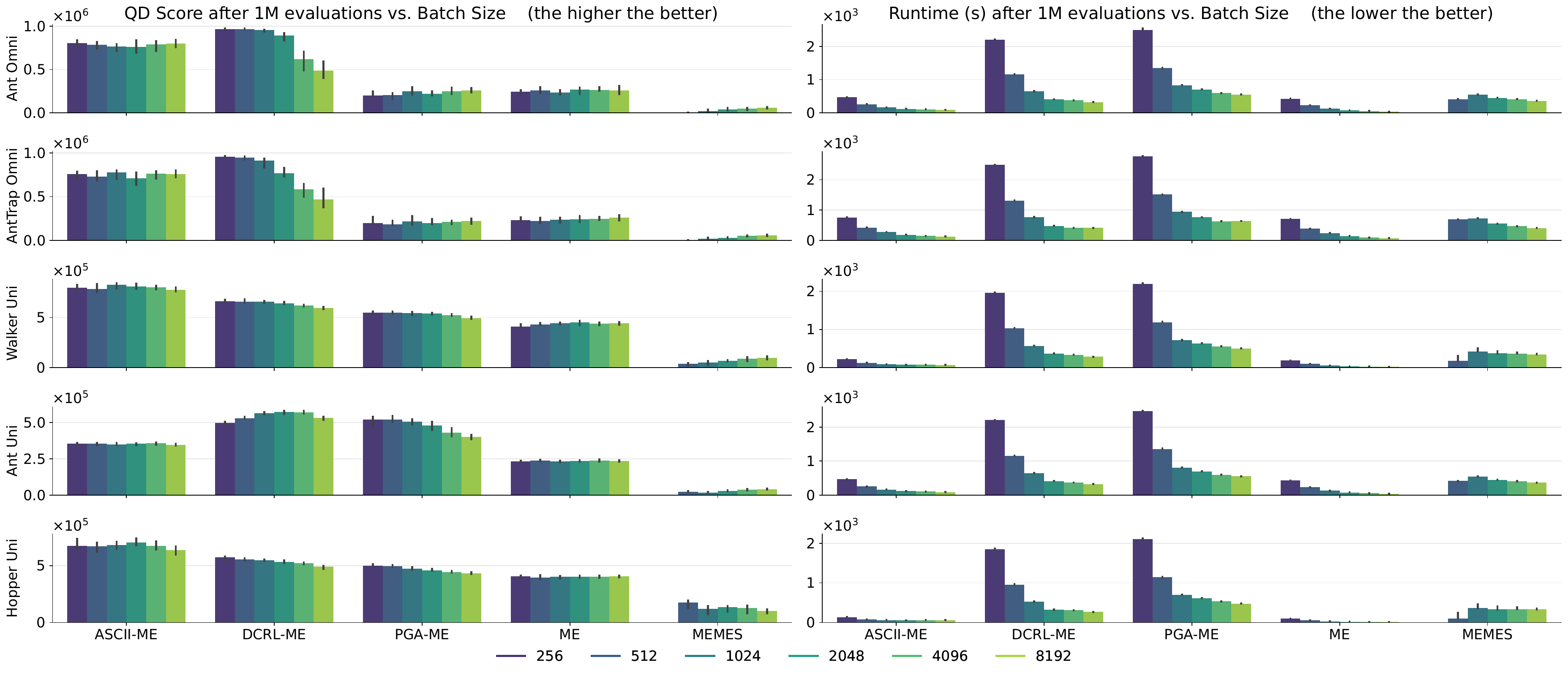}
  \caption{QD score (left) and runtime (right) for \methodName{} and all baselines with varying batch sizes across tasks, after one million evaluations. Vertical lines on bars show lower and upper quartiles; bar height indicates the median over 20 seeds.}
  \label{fig:question2}
\end{figure*}
We consider three main metrics for evaluation: 1) The \textbf{QD-Score} which is the total fitness of all solutions in the archive. In the task at hand, all fitness values are positive, preventing the penalization of algorithms for identifying more solutions. This score reflects both the quality and diversity of the population. 2) The \textbf{Coverage} which is the percentage of occupied cells in the archive, demonstrating coverage of the BD space. 3) The \textbf{Maximum Fitness} which is the fitness of the highest-performing solution in the archive. 

Using the L40S (48GB) GPU, we replicate each experiment 20 times using random seeds and we report p-values based on the Wilcoxon–Mann–Whitney $U$ test with Holm-Bonferroni correction for the quantitative results. The source code for \methodName{} will be made available upon acceptance.

\subsection{Evaluation Procedure}
\label{subsec:efficiency_score}
Increasing the batch size in an algorithm while maintaining a constant evaluation budget can significantly reduce runtime. However, this increase results in fewer iterations, potentially deteriorating the final performance. To fairly evaluate methods in terms of sample and runtime efficiency, we identify the optimal batch size for each algorithm after one million evaluations. This optimal size aims to achieve a balance between low runtime and high final performance, with all other hyperparameters set as recommended by the original authors (see Appendix~\ref{apdx:hyperparameters}). The results presented in the main experiments (see Section~\ref{results:efficiency}), highlight the configurations that best balance performance and runtime. We also detail the methodology used to select these configurations and analyze the impact of batch size variations on each algorithm (see Section~\ref{results:scalability}). 

Note that, in addition to the common ME evaluation cycle, PPGA incorporates a PPO evaluation cycle utilizing parallel environments for efficient training. To find the optimal configuration for PPGA, we tune the number of environments used in parallel to maintain a constant update-to-data ratio, similar to the original setup. Due to structural differences, PPGA is omitted from the scalability analysis in Section~\ref{results:scalability} to prevent unfair comparisons.

\subsection{Results \& Analysis}
\subsubsection{\textbf{Sample and Runtime Efficiency}}
\label{results:efficiency}
Figure~\ref{fig:question1} compares the results of \methodName{} against all baselines on all tasks in terms of both sample and runtime efficiency. The first observation is that \methodName{} achieves the highest QD scores among all baselines in the unidirectional tasks both under conditions of equal runtime and equal evaluation usage, with the exception of the Ant Uni task. Specifically, with a budget of one million evaluations, \methodName{} achieves on average $25\%$ higher QD score than the state-of-the-art algorithm, \dcrl{}, while being on average five times faster in the Hopper Uni and Walker Uni tasks ($p<1\times10^{-7}$). A similar trend persists throughout the algorithms' runtime. In the Ant Uni task, \methodName{} outperforms all baselines except for \dcrl{} in achieving higher QD scores when all algorithms are run for the same duration, until at least the first quarter of the run ($p < 5\times10^{-6}$). In terms of sample efficiency, with a budget of one million evaluations, it outperforms all baselines except the \pga{} and \dcrl{} ($p<1\times10^{-7}$).

Remarkably, \methodName{} does not use any descriptor information, unlike \dcrl{}, yet it achieves competitive results in omnidirectional tasks. In these tasks, where the fitness function discourages solution diversity, \methodName{} still demonstrates a coverage of $100\%$. Throughout at least the first third of the omnidirectional tasks runtime, \methodName{} achieves the highest QD scores among all baselines ($p<1\times10^{-6}$), performing up to an average of $358\%$ higher QD score than \dcrl{} ($p<5\times10^{-8}$). It is only towards the end of training that \dcrl{} surpasses \methodName{} ($p<5\times10^{-4}$), with an average of $4\%$. In terms of sample efficiency, \methodName{} achieves higher QD scores than all baselines except \dcrl{}, when using one million evaluations ($p<5\times10^{-7}$).

Regarding coverage, \methodName{} achieves the highest scores with fewer evaluations and shorter runtime than all baselines in all tasks. This robust coverage underscores the efficacy of \methodName{} in exploring the solution space. However, the strong coverage of \methodName{}, along with its relatively lower maximum fitness score in Ant Uni task, highlights its inability to identify exceptionally high-performing solutions. This limitation likely stems from the fact that critics make more accurate performance estimations than our method, impairing its search efficiency in tasks with larger state and action spaces, like Ant Uni.

\subsubsection{\textbf{Increasing Parallelization}}
\label{results:scalability}
\begin{figure*}[ht!]
  \centering
  \includegraphics[width=\textwidth]{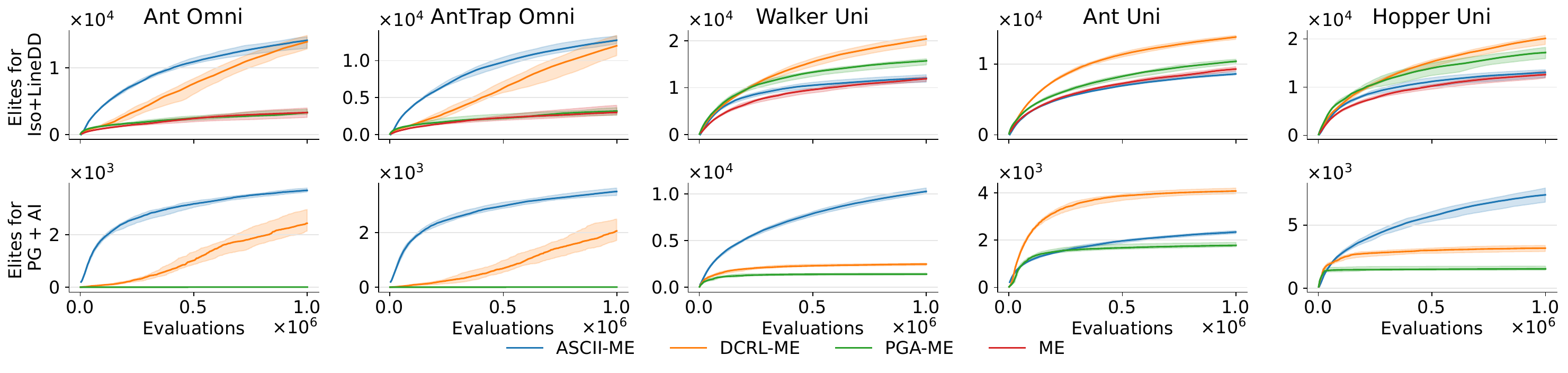}
  \caption{Accumulated number of solutions added to the archive for \isoline{} variation operator and PG variation operator plus Actor Injection (AI). The solid line is the median and the shaded area represents lower and upper quartiles over 20 seeds.}
  \label{fig:elites}
\end{figure*}
\begin{figure}[ht!]
  \centering
  \includegraphics[width=\columnwidth]{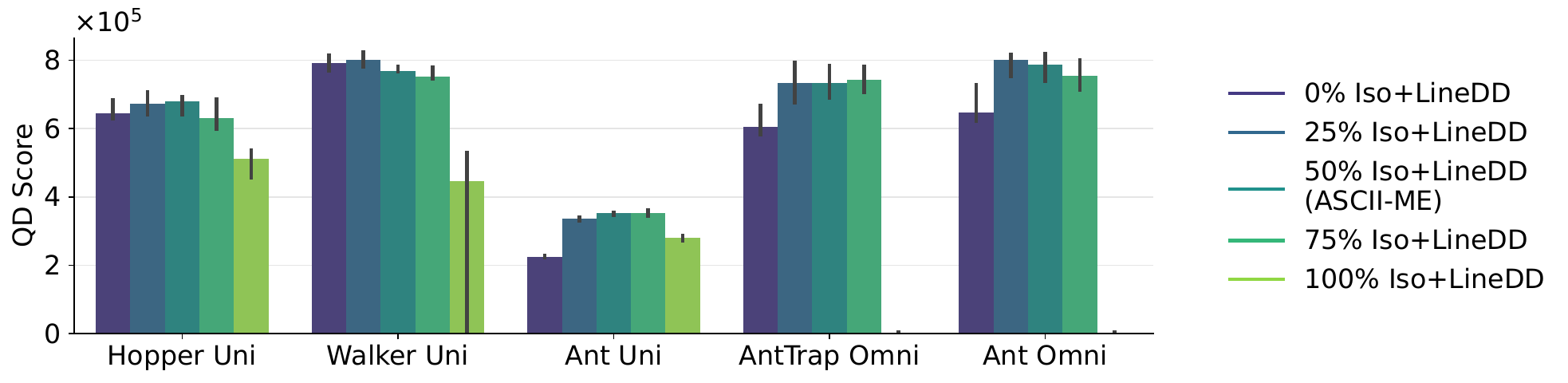}
  \caption{QD score of \methodName{} with different mutation proportions, after one million evaluations. Vertical lines on bars show lower and upper quartiles; bar height indicates the median over 20 seeds.}
  \label{fig:GA_ablation}
\end{figure}
To determine the optimal batch size for each algorithm for use in the experiments in Section~\ref{results:efficiency}, we establish an “efficiency score” for each batch size by following these steps: 1) Perform min-max normalization on QD scores and runtimes for the varying batch sizes used by each algorithm in each task. 2) Adjust runtime scores by subtracting them from 1, ensuring higher values of both QD score and adjusted runtime score indicate better performance. 3) Compute the efficiency score for the varying batch sizes used by each algorithm in each task by multiplying the normalized QD score with the adjusted normalized runtime score. 4) Calculate the mean efficiency score for the varying batch sizes used by each algorithm across all tasks. 5) Select the batch size that yields the highest final efficiency score for each algorithm (see Appendix~\ref{apdx:efficiency_scores}).

Figure~\ref{fig:question2} illustrates the impact of varying batch sizes on the final performance and runtime of each algorithm across all tasks after one million evaluations. Increasing the batch size generally reduces runtime for all algorithms; however, the rate of reduction diminishes with larger sizes due to the constraints of parallel evaluations on the considered GPU. Despite these changes in batch size, the performance of \methodName{} and ME remains remarkably stable, with a mean coefficient of variation (CV) of $2\%$ and $3\%$ across all tasks, respectively. In contrast, the AC-based QD algorithms, \dcrl{} and \pga{}, show an average decline in performance as batch sizes increase, with mean CVs of $13\%$ and $8\%$ across all tasks, respectively. MEMES is the only algorithm that demonstrates an overall positive trend in performance with increasing batch sizes, though it achieves relatively smaller reductions in runtime, primarily due to its higher computational demands.

Focusing on the PG-based algorithms, we observe significant differences in how batch sizes influence QD scores between methods using centralized AC training and \methodName{}, which does not. As outlined in Section~\ref{subsec:ac_based_methods}, in \dcrl{} and \pga{}, AC training utilizes data from evaluations stored in a buffer during each iteration. Although data availability increases in each iteration, the total number of iterations decreases as batch sizes grow. This dynamic raises three strategic questions regarding the AC training: 1) Should the mini-batch size per gradient step be increased to accommodate larger batch sizes? 2) Should the number of critic gradient steps be adjusted to match the total number done with smaller batch sizes? 3) Should the training parameters remain unchanged? These strategies were implemented and tested across three of the five tasks. Based on the ``efficiency score'', strategy 3 proved to be the most efficient for both \dcrl{} and \pga{} (see Appendix~\ref{apdx:ac_strategies}). While strategies 1 and 2 achieved higher QD scores, they did not reduce runtime; in fact, they often resulted in longer runtimes, leading to lower efficiency scores. Using smaller batch sizes across more iterations, with fewer critic gradient updates per iteration, enhances AC training by promoting incremental learning and iterative policy refinement. This strategy mitigates overfitting in individual iterations and improves the critic’s accuracy in policy evaluation, leading to significantly better overall performance of AC-based QD algorithms. Although extensive task-specific hyperparameter tuning could potentially enhance scalability for these algorithms, \methodName{} offers straightforward scalability. With \methodName{}, users can scale up through parallelization without needing to adjust parameters, thus benefiting from reduced runtime and consistently high performance.

\subsubsection{\textbf{Synergizing \variationName{} with \isoline{}}}
To test the synergy between the operators \variationName{} and \isoline{}, we include the QD score of variants of \methodName{} after one million evaluations, varying the proportions of solutions mutated by \isoline{} across all tasks (see Figure~\ref{fig:GA_ablation}). In this setup, $n\%$ \isoline{} corresponds to $(100-n)\%$ \variationName{}.  Our findings indicate that on average, neither $0\%$ nor $100\%$ \isoline{} provides superior performance compared to the mixed variants, indicating the necessity of using both operators. Interestingly, in the omnidirectional tasks, no solutions can be found when our operator \variationName{} is not used. \methodName{}, which includes $50\%$ of \isoline{}, achieves performance similar to both $25\%$ and $75\%$ \isoline{} and is highly robust across all tasks. 

Moreover, we track the number of solutions added to the archive when mutated by each operator for all algorithms that use a PG-based operator and the \isoline{} operator, aiming to understand each operator’s contribution to archive improvement (see Figure~\ref{fig:elites}). We observe that, except in Ant Uni task, the contribution of \variationName{} to the archive is consistently higher than that of other PG-based operators. This indicates the robustness of \variationName{} alone at diversifying and improving the quality of the archive. Additionally, the contribution of \isoline{} in \methodName{} is consistently higher than in ME for omnidirectional tasks. This suggests that \variationName{} effectively navigates the search space to regions where solutions are novel or high-performing, thereby enabling \isoline{} to make variations in these regions and improve the archive. This also suggests that \isoline{} has the potential to contribute more to the archive than it does in ME and to achieve better performances. It only requires a ``navigation'' mechanism that identifies a few novel and high-performing solutions in the search space, allowing \isoline{} to make these variations.

\section{Conclusion}
In this work, we present the \methodName{} algorithm, an enhancement of MAP-Elites that utilizes a policy gradient based operator, \variationName{}, to evolve large deep neural network policies with strong sample and runtime efficiency. Inspired by \isoline{}, this operator interpolates between two action sequences, using time step performance comparisons to generate a new desired action sequence. In contrast to \isoline{}, \variationName{} employs the MDP framework to efficiently bias variations in the action space and uses policy gradients to map these desired action-based changes onto solution parameters. Although \methodName{} does not rely on actor-critic training, it remains competitive with state-of-the-art actor-critic based QD algorithms. Unlike these algorithms, it can straightforwardly scale with massive parallelization, benefiting from lower runtimes without compromising performance on a single consumer-grade GPU. To the best of our knowledge, \methodName{} is the first policy gradient based QD algorithm to operate independently of actor-critic methods. The results showcase a promising framework that sets the stage for future advancements in neuroevolution research using QD.

\bibliographystyle{ACM-Reference-Format}
\bibliography{ref}

\clearpage
\appendix
\section{Supplementary Materials}
\subsection{\methodName{} Pseudocode}
\label{apdx:pseudocode}
In this section, we provide the pseudocode for the \methodName{} algorithm (see Algorithm~\ref{alg:ascii-me}), following the notation introduced and explained in Section~\ref{sec:methods}.
\begin{algorithm}[H]
\caption{\methodName{}}
\label{alg:ascii-me}
\begin{algorithmic}
\small
\Require batch size $k$, number of GA variations $g \leq k$
\State Initialize archive $\mathcal{X}$ with $k$ random solutions and buffer $\mathcal{B}$
\State $n \gets 0$
\While{$n < N$}
    \For{$i=1\rightarrow k$}
    \If{$i \leq g$}
    \State $\bm{x}_i, \bm{x}_{i+1} \gets \textsc{selection}(\mathcal{X})$ \Comment{uniform sampling}
    \State $\hat{\bm{x}}_i \gets \textsc{\isoline{}}(\bm{x}_i, \bm{x}_{i+1})$
    \Else
    \State $\bm{x}_i \gets \textsc{selection}(\mathcal{X})$ \Comment{uniform sampling}
    \State $\hat{\bm{x}}_{i}\gets \textsc{\variationName{}}(\bm{x}_{i}, \mathcal{B})$
    \EndIf
    \State $\textsc{addition}(\hat{\bm{x}}_i, \mathcal{X}, \mathcal{B})$
    \EndFor
    \State $n\gets n + k$  
\EndWhile

\Function{\textsc{addition}}{$\mathcal{X}, \mathcal{B}, \hat{\bm{x}}_i$}
        \State $(f, \text{trajectory}_{\hat{\bm{x}}_i}) \gets F(\hat{\bm{x}}_i)$, $d \gets D(\hat{\bm{x}}_i)$
        \State $\left\{G_t^{i}\right\}_{t=0}^{H-1} \gets \textsc{compute\_rewards-to-go}(\text{trajectory}_{\hat{\bm{x}}_i})$ 
        \State $\text{trajectory}_{\hat{\bm{x}}_i} \gets \textsc{replace}(\text{trajectory}_{\hat{\bm{x}}_i},\left\{r_t^{i}\right\}_{t=0}^{H-1}, \left\{G_t^{i}\right\}_{t=0}^{H-1})$
        \State $\textsc{insert}(\mathcal{B}, \text{trajectory}_{\hat{\bm{x}}_i})$ 
        \If{$\mathcal{X}(d) = \emptyset$ or $F(\mathcal{X}(d)) < f$}
            \State $\mathcal{X}(d) \gets \left\{\hat{\bm{x}}_i,\left\{G_t^{i}\right\}_{t=0}^{H-1},\left\{s_t^{i}\right\}_{t=0}^{H-1}\right\}$
        \EndIf
\EndFunction
\Function{\textsc{\isoline{}}}{$\bm{x}_i,\bm{x}_j$}
    \State $\hat{\bm{x}}_i = \bm{x}_i + \sigma_1 \mathcal{N}(\bm{0},\bm{I}) + \sigma_2 (\bm{x}_j - \bm{x}_i) \mathcal{N}(0,1)$
    \State \Return $\hat{\bm{x}}_i$
\EndFunction
\Function{\textsc{\variationName{}}}{$\bm{x}_i,\mathcal{B}$}
\State $\text{trajectory}_{\bm{\bm{x}_j}} \gets \textsc{selection}(\mathcal{B})$ \Comment{uniform sampling}
\For{$n=1 \rightarrow e$}
\State $\left\{\tilde{\bm{a}}_t^i\right\}_{t=0}^{H-1} \gets \left\{\bm{\mu}_{\bm{x}_i}(\bm{s}_t^{j})\right\}_{t=0}^{H-1}$
\State $\Delta \bm{x}_i \gets \lambda_2 \sum_{t=0}^{H-1} \bm{z}_t(\bm{x}_i, \bm{x}_j) (\bm{a}_t^j - \tilde{\bm{a}}_t^i) \nabla_{\bm{x}} \bm{\mu}_{\bm{x}}(\bm{s}_t^j) \big|_{\bm{x}=\bm{x}_i}$
\State $\bm{x}_i \gets \bm{x}_i +  \Delta{\bm{x}_i}$
\EndFor
\State $\hat{\bm{x}_i} \gets \bm{x}_i$
\State \Return $\hat{\bm{x}}_i$
\EndFunction
\end{algorithmic}
\end{algorithm}

\subsection{Data sampling: Buffer vs. Archive}
\label{apdx:data_sampling}
\begin{figure}[h!]
  \centering
  \includegraphics[width=\columnwidth]{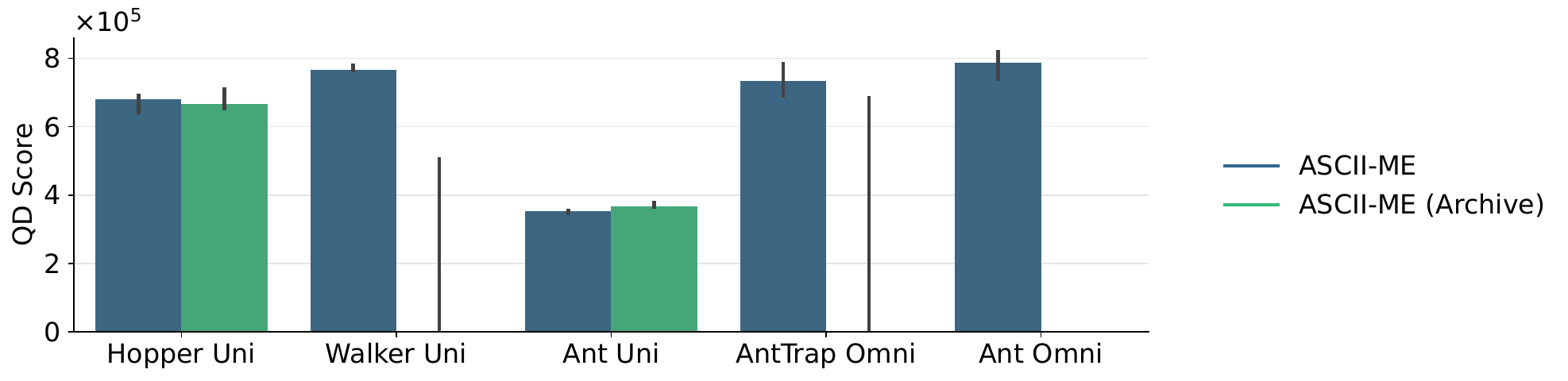}
  \caption{QD score of \methodName{} and \methodName{} (Archive) after one million evaluations. Vertical lines on the bars indicate the lower and upper quartiles; the height of the bars represents the median over 5 seeds.}
  \label{fig:archive_vs_buffer}
\end{figure}

Figure~\ref{fig:archive_vs_buffer} compares the final QD score between \methodName{} and its variant (\methodName{} (Archive)) after one million evaluations. \methodName{} mutates genotypes by sampling trajectories from a buffer, which stores information from previous evaluations of mutated genotypes, regardless of whether they were added to the archive. Consequently, this information may be associated with low-performing genotypes. In contrast, \methodName{} (Archive) samples from the archive, where information is linked to high-performing genotypes. Interestingly, as shown in Figure~\ref{fig:archive_vs_buffer}, \methodName{} (Archive) performs worse, often failing to find any solutions for addition to the archive in three of the five tasks 

\newpage

\subsection{Actor-Critic training Strategies}
\begin{figure*}[ht!]
  \centering
  \includegraphics[width=\textwidth]{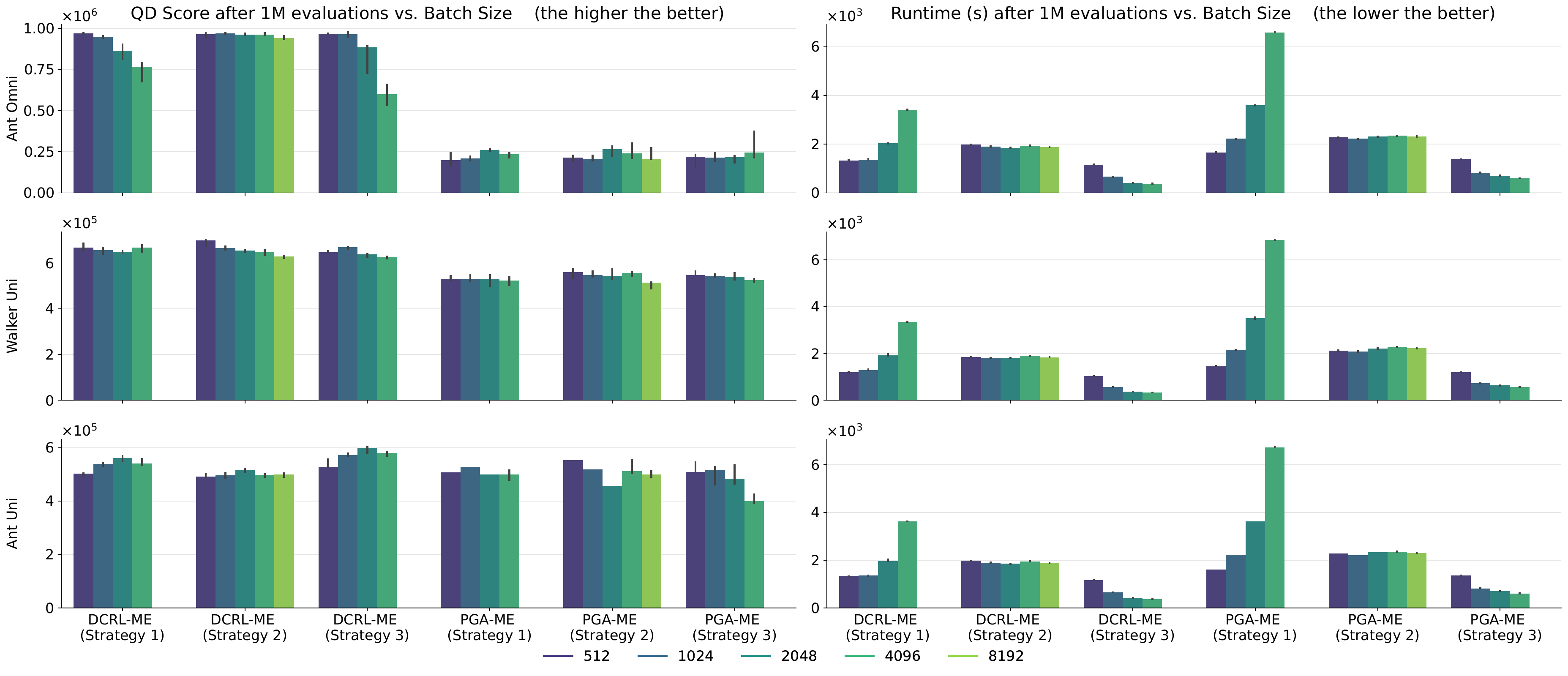}
  \caption{QD score (left) and runtime (right) of different actor-critic strategies with varying batch sizes across tasks, after one million evaluations. Vertical lines on bars show lower and upper quartiles; bar height indicates the median over 5 seeds.}
  \label{fig:ac_strategies}
\end{figure*}

We employ three strategies to investigate the necessary adjustments to the actor-critic training hyperparameters in \dcrl{} and \pga{} with increasing batch sizes. The original hyperparameters, as documented by the authors, are highlighted in bold in Table~\ref{tab:ac_params}. Strategy 1 maintains a constant number of critic training steps across different batch sizes and adjusts the mini-batch size proportionally to the increases in batch sizes. Conversely, Strategy 2 holds the mini-batch size constant while varying the number of critic training steps proportionally with batch size changes. Strategy 3 keeps all actor-critic related hyperparameters unchanged regardless of batch size variations (refer to Table~\ref{tab:ac_params}). Figure~\ref{fig:ac_strategies} demonstrates how the QD score and runtime changes with varying batch size for each strategy when evaluated for one million evaluations. Some results associated with specific batch sizes and strategies are not shown because the computational demands exceeded the capabilities of the available hardware.
\begin{table}
  \caption{Hyperparameters of Actor-Critic Strategies}
  \label{tab:ac_params}
  \begin{tabular}{l l | c c }
    \toprule
    Strategy & Batch Size & mini-batch size & num critic training steps \\
    \midrule
     & \textbf{256} & \textbf{100} & \textbf{3000} \\
     \hline
     1 & 512 & 200 & 3000 \\
      & 1024 & 400 & 3000 \\
      & 2048 & 800 & 3000 \\
      & 4096 & 1600 & 3000 \\
      & 8192 & 3200 & 3000 \\
     \hline
     2 & 512 & 100 & 6000 \\
      & 1024 & 100 & 12000 \\
      & 2048 & 100 & 24000 \\
      & 4096 & 100 & 48000 \\
      & 8912 & 100 & 96000 \\
     \hline
     3 & 512 & 100 & 3000 \\
      & 1024 & 100 & 3000 \\
      & 2048 & 100 & 3000 \\
      & 4096 & 100 & 3000 \\
      & 8912 & 100 & 3000 \\
  \bottomrule
\end{tabular}
\end{table}
\label{apdx:ac_strategies}

\subsection{Hyperparameters of Baseline Algorithms}
\label{apdx:hyperparameters}
Table~\ref{tab:hyperparameters} provides the hyperparameters used for \methodName{} and all baselines across all tasks. The predefined constant of \methodName{}, $\lambda_2$, mentioned in Section~\ref{subsec:variation_eq}, is set to $\lambda_2 =  \frac{\alpha}{H\sigma^{2}}$.
\section{Efficiency Scores}
Table~\ref{tab:batch_sizes} presents the mean ``efficiency score'' across all tasks for each batch size used in each algorithm, which aims to balance high performance and low runtime.
\label{apdx:efficiency_scores}
\textsc{\begin{table}
  \caption{Mean efficiency score across all tasks for each batch size used in each algorithm.}
  \label{tab:batch_sizes}
  \begin{tabular}{l | c c c c c c c c c}
    \toprule
    Algorithm & \multicolumn{9}{c}{Batch Size} \\ 
    \cmidrule(lr){2-10}
              & 256 & 512 & 1024 & 2048 & 3000 & 4096 & 4800 & 6000 & 8129 \\
    \midrule
    \methodName{} & 0.02 & 0.31 & 0.47 & 0.50 &  & \textbf{0.55} &  &  & 0.49\\
    \dcrl{} & 0.02 & 0.42 & 0.64 & \textbf{0.65} &  & 0.48 &  &  & 0.28\\
    \pga{} & 0.02 & 0.30 & \textbf{0.43} & 0.39 &  & 0.35 &  &  & 0.28\\
    \me{} & 0.02 & 0.10 & 0.35 & 0.46 &  & 0.49 &  &  & \textbf{0.58}\\
    MEMES &  & 0.25 & 0.10 & 0.46 &  & 0.49 &  &  & \textbf{0.51}\\
    PPGA &  &  &  &  & 0.04 &  & 0.37 & \textbf{0.46} & \\
  \bottomrule
\end{tabular}
\end{table}}
\textsc{\begin{table*}
  \caption{Hyperparameters for \methodName{} and all baselines}
  \label{tab:hyperparameters}
  \begin{tabular}{l | c c c c c c}
    \toprule
    Parameter & MAP-Elites & MEMES & PGA-ME & DCRL-ME & PPGA & \methodName{}\\
    \midrule
    Number of centroids & $1024$ & $1024$ & $1024$ & $1024$ & $1024$ & $1024$\\
    Evaluation batch size $k$ & $8192$ & $8192$ & $1024$ & $2048$ & $300$ & $4096$\\
    Policy networks & [64, 64, $|\mathcal{A}|$] & [64, 64, $|\mathcal{A}|$] & [64, 64, $|\mathcal{A}|$] & [64, 64, $|\mathcal{A}|$] & [64, 64, $|\mathcal{A}|$] & [64, 64, $|\mathcal{A}|$]\\
    GA batch size $k_{GA}$ & 8192 & 0 & 512 & 1024 & & 2048\\
    PG batch size $k_{PG}$ & 0 & 0 & 511 & 512 & & 2048\\
    AI batch size $k_{AI}$ & 0 & 0 & 1 & 512 & & 0\\
    \hline
    GA variation param. 1 $\sigma_1$ & $0.005$ &  & $0.005$ & $0.005$ & & $0.005$\\
    GA variation param. 2 $\sigma_2$ & $0.05$ &  & $0.05$ & $0.05$ & & $0.05$\\
    \hline
    Actor network &  &  & [64, 64, $|\mathcal{A}|$] & [64, 64, $|\mathcal{A}|$] & [64, 64, $|\mathcal{A}|$] & \\
    Critic network &  &  & [256, 256, 1] & [256, 256, 1] & [256, 256, 1] & \\
    TD3 batch size $N$ &  &  & $100$ & $100$ & & \\
    Critic training steps  &  &  & $3000$ & $3000$ & & \\
    PG training steps $e$ &  &  & $150$ & $150$ & & $32$\\
    Policy learning rate $\alpha$ &  &  & $5 \times 10^{-3}$ & $5 \times 10^{-3}$ & $1 \times 10^{-3}$ & $3 \times 10^{-3}$\\
    Actor learning rate &  &  & $3 \times 10^{-4}$ & $3 \times 10^{-4}$ & & \\
    Critic learning rate &  &  & $3 \times 10^{-4}$ & $3 \times 10^{-4}$ & & \\
    Buffer size &  &  & $4 \times 10^6$ & $8 \times 10^6$ & & $1.024 \times 10^6$\\
    Discount factor $\gamma$ &  &  & $0.99$ & $0.99$ & & $0.99$\\
    Actor delay &  &  & $2$ & $2$ & & \\
    Target update rate &  &  & $0.005$ & $0.005$ & & \\
    Policy noise  &  &  & $0.2$ & $0.2$ & & \\
    Smoothing noise clip &  &  & $0.5$ & $0.5$ & & \\
    Reward scaling &  &  & $1$ & $1$ & & \\
    Length scale $l$ &  &  &  & 0.1 & & \\
    Noise variance $\sigma^{2}$ &  &  & & & & $4$ \\
    Clipping ratio $\epsilon$ &  &  & & & & $0.8$ \\
    Cosine min threshold  $b$ &  &  & & & & $0.25$ \\
    \hline
    Sample number &  & 1000 &  &  & &\\
    Sample sigma &  & 0.02 &  &  & &\\
    \hline
    Num minibatches (PPO) &  &  &  &   & 8 &\\
    Num epochs (PPO) &  &  &  &   & 4 &\\
    Env batch size (PPO) &  &  &  &   & 6000 &\\
    Rollout length (PPO)&  &  &  &   & 64 &\\
    Value function coefficient (PPO) &  &  &  &   & 2& \\
    Max gradient norm (PPO) &  &  &  &   & 1& \\
    Normalize obs (PPO) &  &  &  &   & true& \\
    Normalize returns (PPO) &  &  &  &   & true& \\
    \hline
    Calc gradient iters &  &  &  &   & 10& \\
    Move mean iters &  &  &  &   & 10& \\
    Archive learning rate &  &  &  &   & $\{0.1, 0.5\}$& \\
    Threshold min &  &  &  &   & $\{-500, 200\}$& \\
    $\sigma_0$ &  &  &  &   & $\{1, 3\}$& \\
  \bottomrule
\end{tabular}
\end{table*}}

\end{document}